\documentclass[conference]{IEEEtran}
\IEEEoverridecommandlockouts

\usepackage[utf8]{inputenc}
\usepackage[T1]{fontenc}
\usepackage{microtype,newtxtext,newtxmath} 
\usepackage{mathtools,gensymb,eurosym,url,soul} 
\usepackage[font=footnotesize,textfont=footnotesize]{caption,subcaption}
\usepackage{longtable,booktabs,array}
\usepackage[backend=biber, sorting=none, style=ieee, maxnames=5]{biblatex}
\usepackage[switch]{lineno}
\usepackage{acronym}
\usepackage{graphicx}
\usepackage{hyperref}
\hypersetup{colorlinks=true, breaklinks=true, 
  urlcolor=blue, linkcolor=blue,anchorcolor=blue,citecolor=blue,
  hypertexnames=true, final=true,
  pdfpagemode = UseNone, 
  pdfauthor = {},
  pdftitle = {},
  pdfsubject = {},
  pdfkeywords = {}
}

\graphicspath{{figures/} {img/} {./}}
\DeclareGraphicsExtensions{.pdf,.png,.jpg,.mps}

\def\BibTeX{{\rm B\kern-.05em{\sc i\kern-.025em b}\kern-.08em
    T\kern-.1667em\lower.7ex\hbox{E}\kern-.125emX}}
\DeclareSourcemap{
  \maps[datatype=bibtex]{
    \map[overwrite=true]{
      \step[fieldset=url, null]
      \step[fieldset=issn, null]
      \step[fieldset=doi, null]
      \step[fieldset=eprint, null]
    }
  }
}

\IEEEdisplaynontitleabstractindextext

\acrodef{ADC}[ADC]{Analog-to-Digital Converter}
\acrodef{ADEXP}[AdExp-IF]{Adaptive Exponential Integrate-and-Fire}
\acrodef{ADM}[ADM]{Asynchronous Delta Modulator}
\acrodef{AER}[AER]{Address-Event Representation}
\acrodef{AEX}[AEX]{AER EXtension board}
\acrodef{AE}[AE]{Address-Event}
\acrodef{AFE}[AFE]{Analog Front-End}
\acrodef{AFM}[AFM]{Atomic Force Microscope}
\acrodef{AGC}[AGC]{Automatic Gain Control}
\acrodef{AI}[AI]{Artificial Intelligence}
\acrodef{AMDA}[AMDA]{AER Motherboard with D/A converters}
\acrodef{AMPA}[AMPA]{$\alpha$-Amino-3-hydroxy-5-methyl-4-isoxazolepropionic Acid}
\acrodef{ANN}[ANN]{Artificial Neural Network}
\acrodef{API}[API]{Application Programming Interface}
\acrodef{APMOM}[APMOM]{Alternate Polarity Metal On Metal}
\acrodef{ARM}[ARM]{Advanced RISC Machine}
\acrodef{ASIC}[ASIC]{Application Specific Integrated Circuit}
\acrodef{BCM}[BMC]{Bienenstock-Cooper-Munro}
\acrodef{BD}[BD]{Bundled Data}
\acrodef{BEOL}[BEOL]{Back-end of Line}
\acrodef{BG}[BG]{Bias Generator}
\acrodef{BMI}[BMI]{Brain-Machince Interface}
\acrodef{BTB}[BTB]{Band-to-Band tunnelling}
\acrodef{CAD}[CAD]{Computer Aided Design}
\acrodef{CAM}[CAM]{Content Addressable Memory}
\acrodef{CAVIAR}[CAVIAR]{Convolution AER Vision Architecture for Real-Time}
\acrodef{CA}[CA]{Cortical Automaton}
\acrodef{CCN}[CCN]{Cooperative and Competitive Network}
\acrodef{CDR}[CDR]{Clock-Data Recovery}
\acrodef{CFC}[CFC]{Current to Frequency Converter}
\acrodef{CHP}[CHP]{Communicating Hardware Processes}
\acrodef{CMIM}[CMIM]{Metal-Insulator-Metal Capacitor}
\acrodef{CML}[CML]{Current Mode Logic}
\acrodef{CMOL}[CMOL]{Hybrid CMOS nanoelectronic circuits}
\acrodef{CMOS}[CMOS]{Complementary Metal-Oxide-Semiconductor}
\acrodef{CNN}[CNN]{Convolutional Neural Network}
\acrodef{CNS}[CNS]{central Nervous System}
\acrodef{COTS}[COTS]{Commercial Off-The-Shelf}
\acrodef{CPG}[CPG]{Central Pattern Generator}
\acrodef{CPLD}[CPLD]{Complex Programmable Logic Device}
\acrodef{CPU}[CPU]{Central Processing Unit}
\acrodef{CSM}[CSM]{Cortical State Machine}
\acrodef{CSP}[CSP]{Constraint Satisfaction Problem}
\acrodef{CTXCTL}[CTXCTL]{CortexControl}
\acrodef{CV}[CV]{Coefficient of Variation}
\acrodef{DAC}[DAC]{Digital to Analog Converter}
\acrodef{DAS}[DAS]{Dynamic Auditory Sensor}
\acrodef{DAVIS}[DAVIS]{Dynamic and Active Pixel Vision Sensor}
\acrodef{DBN}[DBN]{Deep Belief Network}
\acrodef{DBS}[DBS]{Deep Brain Stimulation}
\acrodef{DFA}[DFA]{Deterministic Finite Automaton}
\acrodef{DIBL}[DIBL]{Drain-Induced Barrier-Lowering}
\acrodef{DI}[DI]{Delay Insensitive}
\acrodef{DMA}[DMA]{Direct Memory Access}
\acrodef{DNF}[DNF]{Dynamic Neural Field}
\acrodef{DNN}[DNN]{Deep Neural Network}
\acrodef{DOF}[DOF]{Degrees of Freedom}
\acrodef{DPE}[DPE]{Dynamic Parameter Estimation}
\acrodef{DPI}[DPI]{Differential Pair Integrator}
\acrodef{DRAM}[DRAM]{Dynamic Random Access Memory}
\acrodef{DRRZ}[DR-RZ]{Dual-Rail Return-to-Zero}
\acrodef{DR}[DR]{Dual Rail}
\acrodef{DSP}[DSP]{Digital Signal Processor}
\acrodef{DVS}[DVS]{Dynamic Vision Sensor}
\acrodef{DYNAP}[DYNAP]{Dynamic Neuromorphic Asynchronous Processor}
\acrodef{EBL}[EBL]{Electron Beam Lithography}
\acrodef{ECG}[ECG]{Electrocardiography}
\acrodef{ECoG}[ECoG]{Electrocorticography}
\acrodef{EDVAC}[EDVAC]{Electronic Discrete Variable Automatic Computer}
\acrodef{EEG}[EEG]{Electroencephalography}
\acrodef{EIN}[EIN]{Excitatory-Inhibitory Network}
\acrodef{EMG}[EMG]{Electromyography}
\acrodef{EM}[EM]{Expectation Maximization}
\acrodef{EOG}[EOG]{Electrooculogram}
\acrodef{EPSC}[EPSC]{Excitatory Post-Synaptic Current}
\acrodef{EPSP}[EPSP]{Excitatory Post-Synaptic Potential}
\acrodef{EZ}[EZ]{Epileptogenic Zone}
\acrodef{FDSOI}[FDSOI]{Fully-Depleted Silicon on Insulator}
\acrodef{FET}[FET]{Field-Effect Transistor}
\acrodef{FFT}[FFT]{Fast Fourier Transform}
\acrodef{FI}[F-I]{Frequency--Current}
\acrodef{FMA}[FMA]{Floating Microelectrode Array}
\acrodef{FNN}[FNN]{Feed-forward Neural Network}
\acrodef{FPGA}[FPGA]{Field Programmable Gate Array}
\acrodef{FR}[FR]{Fast Ripple}
\acrodef{FSA}[FSA]{Finite State Automaton}
\acrodef{FSM}[FSM]{Finite State Machine}
\acrodef{GABA}[GABA]{$\gamma$-Aminobutanoic Acid}
\acrodef{GIDL}[GIDL]{Gate-Induced Drain Leakage}
\acrodef{GOPS}[GOPS]{Giga-Operations per Second}
\acrodef{GPIO}[GPIO]{General Purpose I/O}
\acrodef{GPU}[GPU]{Graphical Processing Unit}
\acrodef{GT}[GT]{Ground Truth}
\acrodef{GUI}[GUI]{Graphical User Interface}
\acrodef{HAL}[HAL]{Hardware Abstraction Layer}
\acrodef{HFO}[HFO]{High Frequency Oscillation}
\acrodef{HH}[H\&H]{Hodgkin \& Huxley}
\acrodef{HMM}[HMM]{Hidden Markov Model}
\acrodef{HRS}[HRS]{High-Resistive State}
\acrodef{HR}[HR]{Human Readable}
\acrodef{HSE}[HSE]{Handshaking Expansion}
\acrodef{HW}[HW]{Hardware}
\acrodef{ICT}[ICT]{Information and Communication Technology}
\acrodef{IC}[IC]{Integrated Circuit}
\acrodef{IF2DWTA}[IF2DWTA]{Integrate \& Fire 2-Dimensional WTA}
\acrodef{IFSLWTA}[IFSLWTA]{Integrate \& Fire Stop Learning WTA}
\acrodef{IF}[I\&F]{Integrate-and-Fire}
\acrodef{IMU}[IMU]{Inertial Measurement Unit}
\acrodef{INCF}[INCF]{International Neuroinformatics Coordinating Facility}
\acrodef{INI}[INI]{Institute of Neuroinformatics}
\acrodef{IO}[I/O]{Input/Output}
\acrodef{IPSC}[IPSC]{Inhibitory Post-Synaptic Current}
\acrodef{IPSP}[IPSP]{Inhibitory Post-Synaptic Potential}
\acrodef{IP}[IP]{Intellectual Property}
\acrodef{ISI}[ISI]{Inter-Spike Interval}
\acrodef{IoT}[IoT]{Internet of Things}
\acrodef{JFLAP}[JFLAP]{Java - Formal Languages and Automata Package}
\acrodef{LEDR}[LEDR]{Level-Encoded Dual-Rail}
\acrodef{LFP}[LFP]{Local Field Potential}
\acrodef{LIFE}[LIFE]{Longitudinal Intrafascicular Electrodes}
\acrodef{LIF}[LI\&F]{Leaky Integrate-and-Fire}
\acrodef{LLC}[LLC]{Low Leakage Cell}
\acrodef{LNA}[LNA]{Low-Noise Amplifier}
\acrodef{LPF}[LPF]{Low Pass Filter}
\acrodef{LRS}[LRS]{Low-Resistive State}
\acrodef{LSM}[LSM]{Liquid State Machine}
\acrodef{LTD}[LTD]{Long Term Depression}
\acrodef{LTI}[LTI]{Linear Time-Invariant}
\acrodef{LTP}[LTP]{Long Term Potentiation}
\acrodef{LTU}[LTU]{Linear Threshold Unit}
\acrodef{LUT}[LUT]{Look-Up Table}
\acrodef{LVDS}[LVDS]{Low Voltage Differential Signaling}
\acrodef{MCMC}[MCMC]{Markov-Chain Monte Carlo}
\acrodef{MEA}[MEA]{Multielectrode Arrays}
\acrodef{MEMS}[MEMS]{Micro Electro Mechanical System}
\acrodef{MFR}[MFR]{Mean Firing Rate}
\acrodef{MIM}[MIM]{Metal Insulator Metal}
\acrodef{MLP}[MLP]{Multilayer Perceptron}
\acrodef{ML}[ML]{Machine Learning}
\acrodef{MOSCAP}[MOSCAP]{Metal Oxide Semiconductor Capacitor}
\acrodef{MOSFET}[MOSFET]{Metal Oxide Semiconductor Field-Effect Transistor}
\acrodef{MOS}[MOS]{Metal Oxide Semiconductor}
\acrodef{MRI}[MRI]{Magnetic Resonance Imaging}
\acrodef{NCS}[NCS]{Neuromorphic Cognitive Systems}
\acrodef{NDFSM}[NDFSM]{Non-deterministic Finite State Machine}
\acrodef{ND}[ND]{Noise-Driven}
\acrodef{NEF}[NEF]{Neural Engineering Framework}
\acrodef{NHML}[NHML]{Neuromorphic Hardware Mark-up Language}
\acrodef{NIL}[NIL]{Nano-Imprint Lithography}
\acrodef{NI}[NI]{Neural Interface}
\acrodef{NMDA}[NMDA]{\textit{N}-Methyl-\textsc{d}-aspartate}
\acrodef{NME}[NE]{Neuromorphic Engineering}
\acrodef{NN}[NN]{Neural Network}
\acrodef{NOC}[NoC]{Network-on-Chip}
\acrodef{NRZ}[NRZ]{Non-Return-to-Zero}
\acrodef{NSM}[NSM]{Neural State Machine}
\acrodef{OR}[OR]{Operating Room}
\acrodef{OTA}[OTA]{Operational Transconductance Amplifier}
\acrodef{PCB}[PCB]{Printed Circuit Board}
\acrodef{PCHB}[PCHB]{Pre-Charge Half-Buffer}
\acrodef{PCM}[PCM]{Phase Change Memory}
\acrodef{PC}[PC]{Personal Computer}
\acrodef{PDK}[PDK]{Process Design Kit}
\acrodef{PE}[PE]{Phase Encoding}
\acrodef{PFA}[PFA]{Probabilistic Finite Automaton}
\acrodef{PFC}[PFC]{Prefrontal Cortex}
\acrodef{PFM}[PFM]{Pulse Frequency Modulation}
\acrodef{PNI}[PNI]{Peripheral Nerve Interface}
\acrodef{PNS}[PNS]{Peripheral Nervous System}
\acrodef{PPG}[PPG]{Photoplethysmography}
\acrodef{PR}[PR]{Production Rule}
\acrodef{PSC}[PSC]{Post-Synaptic Current}
\acrodef{PSP}[PSP]{Post-Synaptic Potential}
\acrodef{PSTH}[PSTH]{Peri-Stimulus Time Histogram}
\acrodef{PV}[PV]{Parvalbumin}
\acrodef{QDI}[QDI]{Quasi Delay Insensitive}
\acrodef{RAM}[RAM]{Random Access Memory}
\acrodef{RA}[RA]{Resected Area}
\acrodef{RDF}[RDF]{Random Dopant Fluctuation}
\acrodef{RELU}[ReLu]{Rectified Linear Unit}
\acrodef{RISC}[RISC]{Reduced Instruction Set Computer}
\acrodef{RLS}[RLS]{Recursive Least-Squares}
\acrodef{RMSE}[RMSE]{Root Mean Square-Error}
\acrodef{RMS}[RMS]{Root Mean Square}
\acrodef{RNN}[RNN]{Recurrent Neural Network}
\acrodef{ROLLS}[ROLLS]{Reconfigurable On-Line Learning Spiking}
\acrodef{RRAM}[R-RAM]{Resistive Random Access Memory}
\acrodef{RSA}[RSA]{Respiratory Sinus Arrhythmia}
\acrodef{R}[R]{Ripple}
\acrodef{SAC}[SAC]{Selective Attention Chip}
\acrodef{SAT}[SAT]{Boolean Satisfiability Problem}
\acrodef{SCI}[SCI]{Spinal Cord Injury}
\acrodef{SCX}[SCX]{Silicon CorteX}
\acrodef{SD}[SD]{Signal-Driven}
\acrodef{SEM}[SEM]{Spike-based Expectation Maximization}
\acrodef{SLAM}[SLAM]{Simultaneous Localization and Mapping}
\acrodef{SNN}[SNN]{Spiking Neural Network}
\acrodef{SNR}[SNR]{Signal to Noise Ratio}
\acrodef{SOC}[SoC]{System-On-Chip}
\acrodef{SOI}[SOI]{Silicon on Insulator}
\acrodef{SOZ}[SOZ]{Seizure Onset Zone}
\acrodef{SPI}[SPI]{Serial Peripheral Interface}
\acrodef{SP}[SP]{Separation Property}
\acrodef{SRAM}[SRAM]{Static Random Access Memory}
\acrodef{SST}[SST]{Somatostatin}
\acrodef{STDP}[STDP]{Spike-Timing Dependent Plasticity}
\acrodef{STD}[STD]{Short-Term Depression}
\acrodef{STP}[STP]{Short-Term Plasticity}
\acrodef{STT-MRAM}[STT-MRAM]{Spin-Transfer Torque Magnetic Random Access Memory}
\acrodef{STT}[STT]{Spin-Transfer Torque}
\acrodef{SVM}[SVM]{Support Vector Machine}
\acrodef{SW}[SW]{Software}
\acrodef{TCAM}[TCAM]{Ternary Content-Addressable Memory}
\acrodef{TFT}[TFT]{Thin Film Transistor}
\acrodef{TIME}[TIME]{Transverse Intrafascicular Multichannel Electrode}
\acrodef{TLE}[TLE]{Temporal Lobe Epilepsy}
\acrodef{UEA}[UEA]{Utah Electrode Array}
\acrodef{USB}[USB]{Universal Serial Bus}
\acrodef{USEA}[USEA]{Utah Slanted Electrode Array}
\acrodef{VHDL}[VHDL]{VHSIC Hardware Description Language}
\acrodef{VHSIC}[VHSIC]{Very High Speed Integrated Circuits}
\acrodef{VIP}[VIP]{Vasoactive Intestinal Peptide}
\acrodef{VLSI}[VLSI]{Very Large Scale Integration}
\acrodef{VNS}[VNS]{Vagal Nerve Stimulation}
\acrodef{VOR}[VOR]{Vestibulo-Ocular Reflex}
\acrodef{VSA}[VSA]{Vector Symbolic Architecture}
\acrodef{WCST}[WCST]{Wisconsin Card Sorting Test}
\acrodef{WTA}[WTA]{Winner-Take-All}
\acrodef{XML}[XML]{eXtensible Mark-up Language}
\acrodef{divmod3}[DIVMOD3]{Divisibility of a number by three}
\acrodef{hWTA}[hWTA]{Hard Winner-Take-All}
\acrodef{iEEG}[iEEG]{Intracranial Electroencephalography}
\acrodef{rSNN}[rSNN]{recurrent Spiking Neural Network}
\acrodef{sWTA}[sWTA]{soft Winner-Take-All}

\begin{document}

\author{
  \IEEEauthorblockN{
     Dmitrii Zendrikov\IEEEauthorrefmark{1},
     Alessio Franci\IEEEauthorrefmark{2}\IEEEauthorrefmark{4},
     Giacomo Indiveri\IEEEauthorrefmark{1}\IEEEauthorrefmark{4}
  }
   \IEEEauthorblockA{\IEEEauthorrefmark{1}Institute of Neuroinformatics, University of Zurich and ETH Zurich Switzerland}
   \IEEEauthorblockA{\IEEEauthorrefmark{2}University of Liege, Belgium, and WEL Research Institute, Wavre, Belgium.}
}

\title{Waves and symbols in neuromorphic hardware: from analog signal processing to digital computing on the same computational substrate
\thanks{\IEEEauthorrefmark{4}Equally contributing senior authors.}
\thanks{This work was partially supported by the European Research Council (ERC) under the European Union’s Horizon 2020 Research and Innovation Program Grant Agreement No. 724295 (NeuroAgents).}
}

\maketitle

\begin{abstract}
  Neural systems use the same underlying computational substrate to carry out analog filtering and signal processing operations, as well as discrete symbol manipulation and digital computation.
  Inspired by the computational principles of canonical cortical microcircuits, we propose a framework for using recurrent spiking neural networks to seamlessly and robustly switch between analog signal processing and categorical and discrete computation.
  We provide theoretical analysis and practical neural network design tools to formally determine the conditions for inducing this switch.
  We demonstrate the robustness of this framework experimentally with hardware soft Winner-Take-All and mixed-feedback recurrent spiking neural networks, implemented by appropriately configuring the analog neuron and synapse circuits of a mixed-signal neuromorphic processor chip.
\end{abstract}


\section{Introduction}
\label{sec:introduction}

Biological neural processing systems use the same computational substrate to process both analog sensory signals, such as pressure waves, and manipulate symbolic information, such as concepts and categories.
For example, experimental neuroscience studies have demonstrated that animal brains are capable of amplifying, filtering, and normalizing sensory inputs with different contrast and frequency components, as well as categorizing input perceptual patterns into discrete sets of objects using the same cortical circuits~\cite{Douglas_Martin07,Carandini_Heeger12,VanRullen_Koch03}.
Similarly, classic neuroscience experiments on decision making demonstrate how neural circuits can use linear operators, such as integrators and filters, to accumulate evidence and digital operators to produce binary decisions~\cite{Wang22,Mante_etal13}.
In addition, a gradual and smooth transition from analog signal integration to discrete decision making has been observed at increasingly detailed levels of organization extending beyond the neural circuit level, ranging from unicellular organisms~\cite{zeng2010decision} to animal collectives~\cite{couzin2005effective}.

In contrast, in electrical engineering and computer science, there has always been a clear distinction between the domains of ``signal processing'' and ``computing'',  the former typically implemented with analog circuits or \acp{DSP} and the latter with synchronous digital circuits and microprocessors.
To overcome the limitations of such a dichotomy, a novel mixed-feedback framework was recently developed on the ground of bio-inspired control-theory~\cite{sepulchre2019control,leonard2024fast}, providing analysis tools and design principles for regulating the transition between analog signal processing and discrete computation.

Neuromorphic engineering and computing represent a novel field that combines the principles of computation used by biological nervous systems with those of electrical engineering and computer science~\cite{Mead23}, which can best exploit this mixed-feedback framework~\cite{Hahnloser_etal00}.

Here, we show how populations of neurons implemented using mixed-signal analog/digital neuromorphic circuits can be configured using mixed-feedback principles to smoothly
transition from linear processing on distributed input signals to discrete symbolic data selection and manipulation, bridging the gap between signal processing and classical computing.

We provide experimental results that demonstrate how such a robust transition between the two regimes can be achieved and show how this framework provides crucial flexibility for processing incoming signals, with the same substrate, in different ways as a function of context.

\section{Signal processing and information representation in populations of spiking neurons}
\label{sec:sign-proc-inform}

\subsection{Robust representations using inhomogeneous units}
\label{sec:robust-repr-using}

Typically, artificial neural networks use single units with high precision real-value outputs to represent signals and variables.
In contrast, animal brains use low-precision spiking neurons and \emph{population coding} strategies to represent signals robustly~\cite{Averbeck_etal06,Pouget_etal00}.
For example, in primary cortical sensory areas signals are represented via ``feature maps'' comprising populations of recurrently connected excitatory and inhibitory neurons arranged on the surface of the cortex (e.g., orientation maps in the visual cortex, tonotopic maps in the auditory cortex, somatotopic maps in the somatosensory cortex, etc.)~\cite{Bednar_Wilson16}.
The strategy of employing populations of variable units to improve robustness in representing signals is an optimal one also for mixed-signal neuromorphic hardware systems that use analog circuits affected by device mismatch to emulate spiking neurons~\cite{Zendrikov_etal23}.

\begin{figure}
  \begin{subfigure}{0.5\textwidth}
    \centering
    \includegraphics[width=0.75\textwidth]{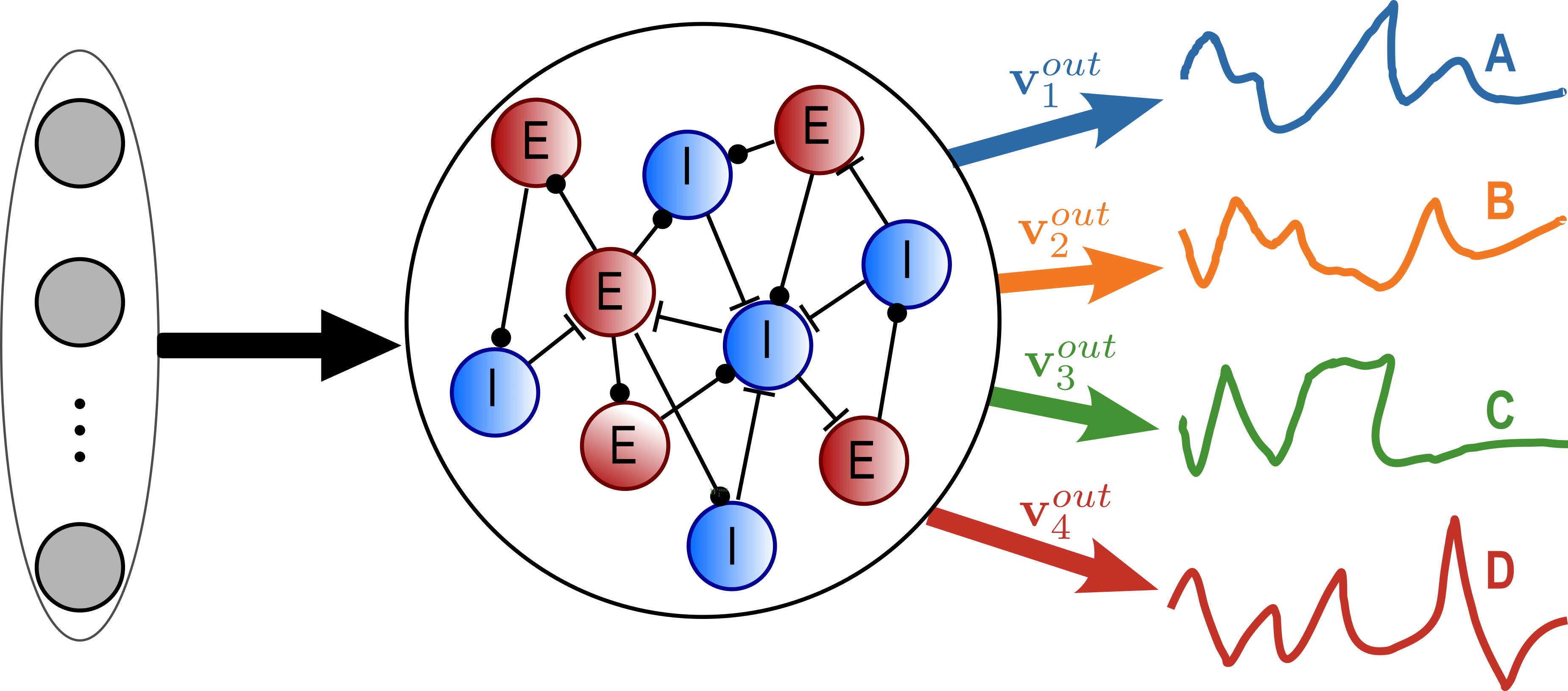}
    \caption{}
    \label{fig:sRNNnet}
  \end{subfigure}
  \begin{subfigure}{0.5\textwidth}
    \centering
    \includegraphics[width=0.75\textwidth]{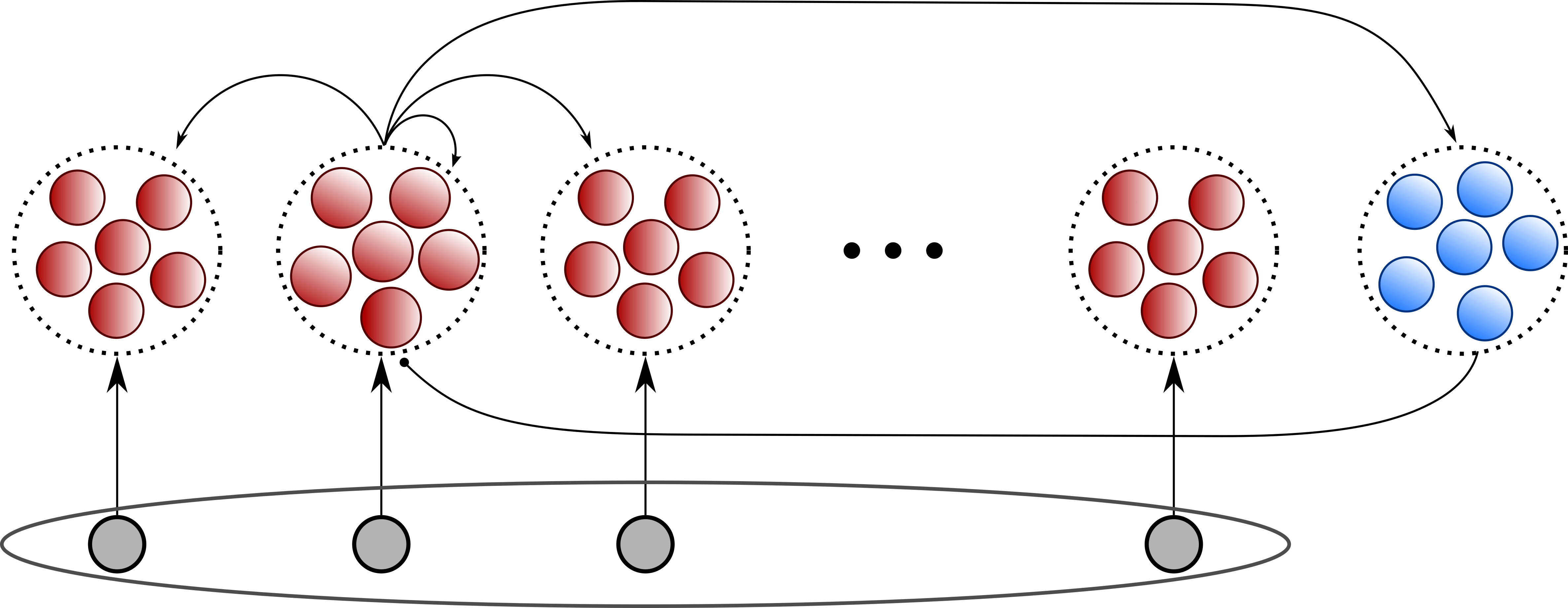}
    \caption{}
    \label{fig:position-wta}
  \end{subfigure}
  \caption{Recurrent Spiking Neural Network (\ac{rSNN}) architectures.
(\subref{fig:sRNNnet}) Generic mixed-feedback \ac{rSNN} with input layer, recurrent pool of excitatory and inhibitory spiking neurons, and readout units that encode continuous time varying outputs; (\subref{fig:position-wta}) Specific \ac{rSNN} architecture in which excitatory and inhibitory neurons have been arranged to form a \ac{sWTA} network.
Recurrent excitatory and inhibitory connections are shown only for one cluster.}
  \label{fig:network_sketch}
\end{figure}

\subsection{Recurrent Spiking Neural Networks}

Figure~\ref{fig:network_sketch} shows examples of recurrent network architectures that support population coding and which can represent promising configurations for achieving robust signal representation in mixed-signal neuromorphic hardware.
The diagram of Fig.~\ref{fig:sRNNnet} shows a generic \ac{rSNN} composed of three layers: an input layer which encodes signals using multiple spiking neurons; a recurrently connected layer of excitatory (E) and inhibitory (I) neurons, for implementing E-I balanced networks; and a read-out layer with units that receive projections from the recurrent network to generate synaptic (low-pass filtered) analog output signals.
These analog signals can be interpreted as the instantaneous strength, or decision, with which the network represents the symbols associated with each output channel (A,B,C,D in Fig.~\ref{fig:sRNNnet}).
The architecture of Fig.~\ref{fig:position-wta} represents a specific instantiation of the generic \ac{rSNN}, configured as a \ac{sWTA}~\cite{Douglas_Martin07}: clusters of excitatory neurons receive common input and excite both their nearest neighbors and a shared pool of inhibitory neurons.
The cluster of inhibitory neurons, in turn, provides global inhibition, stabilizing the activity of the recurrently connected excitatory neurons~\cite{Zendrikov_etal23}.
In these networks, the input layer typically represents a feature map in which nearby neurons encode similar feature properties.
For example, in the case of orientation tuning in visual cortex, nearby neurons would code for similar orientations of oriented visual stimuli.
As demonstrated both in real cortical circuits~\cite{Douglas_Martin07}, and with theoretical analysis formalisms~\cite{Rutishauser_etal11,Rutishauser_etal15} \ac{sWTA} networks can switch from performing linear signal processing operations to non-linear categorization and selection ones, depending on their excitatory and inhibitory gain parameters.

\subsection{\ac{sWTA} network operations on neuromorphic hardware}
\label{sec:acswta-netw-oper}

To validate the theories and models presented in the literature~\cite{Rutishauser_etal11,Rutishauser_etal15}, we implemented a \ac{sWTA} network in neuromorphic hardware, using a multi-core neuromorphic processor comprising four cores of 256 \ac{ADEXP} analog neuron circuits, 64 re-configurable dynamic synapses per neuron, and programmable asynchronous digital routing circuits for (re)configuring the network topology~\cite{Moradi_etal18}.

\begin{figure}
  \begin{subfigure}{0.24\textwidth}
    \includegraphics[width=\textwidth]{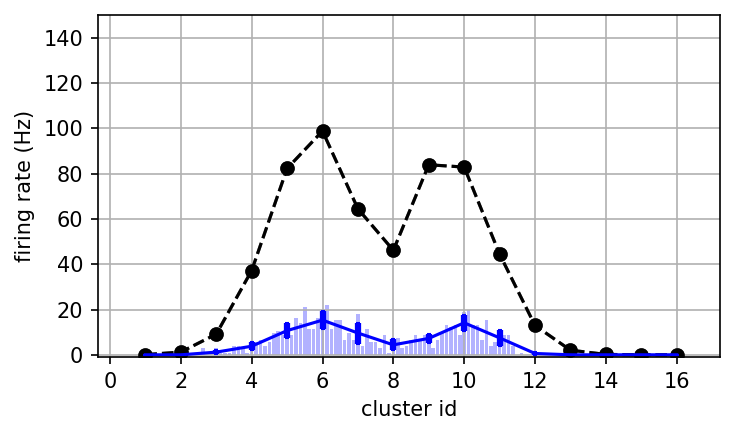}
    \caption{}
    \label{fig:sWTA-2bumps-close-low-alpha}
  \end{subfigure}
  \begin{subfigure}{0.24\textwidth}
    \includegraphics[width=\textwidth]{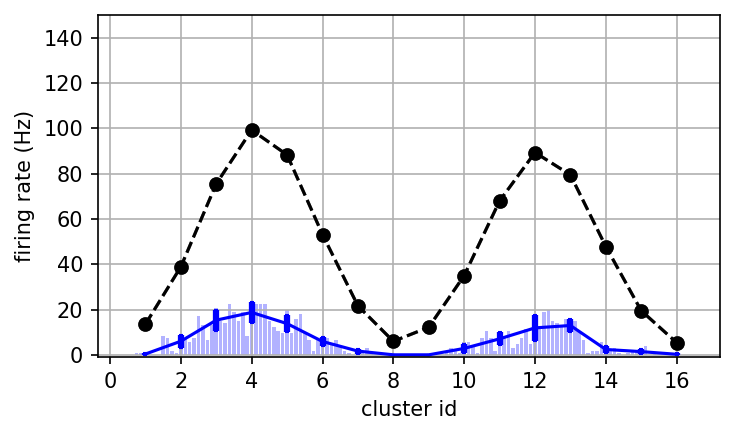}
    \caption{}
    \label{fig:sWTA-2bumps-far-low-alpha}
  \end{subfigure}\\
  \begin{subfigure}{0.24\textwidth}
    \includegraphics[width=\textwidth]{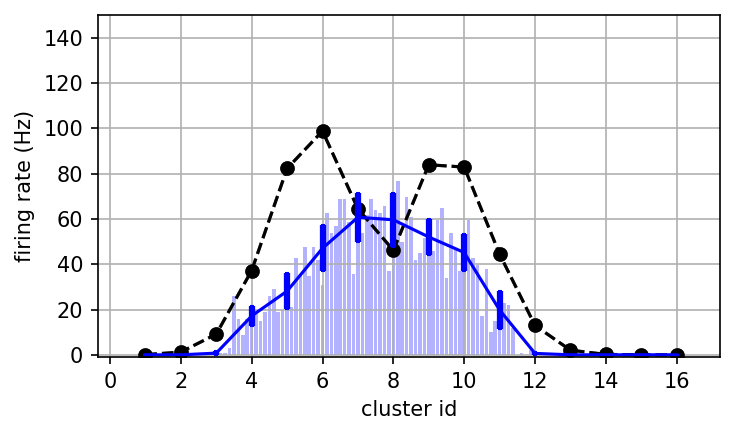}
    \caption{}
    \label{fig:sWTA-2bumps-close-high-alpha}
  \end{subfigure}
  \begin{subfigure}{0.24\textwidth}
    \includegraphics[width=\textwidth]{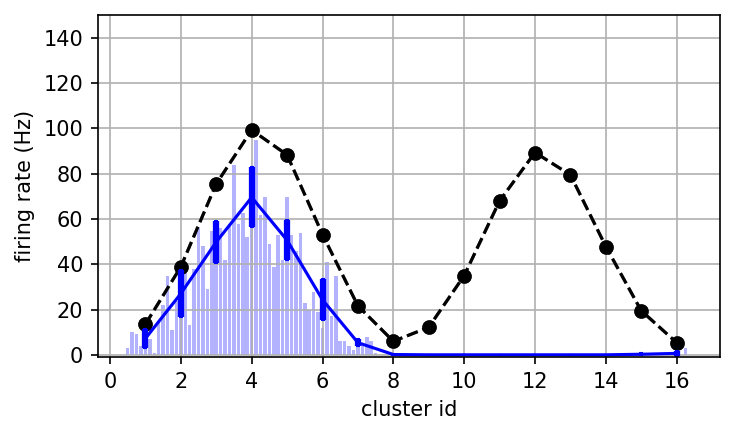}
    \caption{}
    \label{fig:sWTA-2bumps-far-high-alpha}
  \end{subfigure}
  \caption{\ac{sWTA} response properties measured from the neuromorphic chip.
The different input signals are encoded as Poisson spike trains with constant firing rates (dots in black dashed lines).
The mean firing rates of each cluster and their standard deviation are plotted as a solid line with vertical bars in blue.
The shaded bars represent the average firing rate of each neuron in the network (inhibitory neurons not shown).
    In the top row, the network operates in analog signal processing mode.
In the bottom row, the same network with the same inputs but different (higher) gain operates in the categorization mode.}
  \label{fig:sWTA}
\end{figure}

We configured the chip to create a hardware \ac{sWTA} network of 16 excitatory clusters, each comprising eight spiking neurons per cluster, with a structure similar to the one shown in~Fig.~\ref{fig:position-wta}.
The strength of all lateral recurrent excitatory connections is modulated by a global parameter $\alpha$ that sets the gain of the network and controls the transition from analog processing to digital selection~\cite{Hahnloser_etal00,Rutishauser_etal11}.

The network response, measured from the chip, is shown in Fig~\ref{fig:sWTA}.
The input to the network is provided by stimulating each cluster with Poisson spike trains with different mean rates. Figures~\ref{fig:sWTA-2bumps-close-low-alpha} and~\ref{fig:sWTA-2bumps-close-high-alpha} have in input two bumps close in space, while Figures~\ref{fig:sWTA-2bumps-far-low-alpha} and~\ref{fig:sWTA-2bumps-far-high-alpha} have two bumps far apart.

In the top row of Fig.~\ref{fig:sWTA} the network has low values of $\alpha$ and operates in the analog processing mode, while in the bottom row the network has its gain set to high $\alpha$ values, and operates in the categorization mode: if the input bumps are sufficiently close, the network performs signal restoration, and produces a single bump in output; conversely, if the inputs are far apart, the network performs selective amplification and chooses the input with strongest amplitude (the winner), while completely suppressing the one with (even slightly) weaker amplitude.

\subsection{Theoretical analysis of mixed-feedback \acp{rSNN}}
\label{sec:theor-analys-mixed}

The stability and dynamics of \acp{RNN} and \ac{sWTA} architectures have been thoroughly studied analytically in the past.
However, these studies have mainly focused on the mean-rate domain, and few have considered the case of populations of \emph{spiking} neurons.
In particular, rigorously characterizing the input/output behavior of such networks to determine which network structure to use for which use case still remains an open challenge.
To solve this challenge and make the theoretical analysis of the types of \acp{rSNN} depicted in Fig.~\ref{fig:sRNNnet} tractable, we now consider architectures that do not necessarily adhere to
Dale's principle~\cite{Strata_Harvey99}, \emph{i.e.}, networks in which spiking neurons can both excite and inhibit their target neurons.
In this case, the network will be characterized by a connectivity/adjacency matrix $A$ which can have both positive and negative values.
In addition, we restrict our analysis to \emph{mixed-feedback} networks that combine local negative feedback mechanisms with distributed (networked) positive ones.
Specifically, we consider spiking neural networks in which each neuron has two sources of \emph{local negative feedback}: (1) the passive leaky dynamics that regulate the neuron to its resting state, and (2) the spike-frequency adaptation dynamics that regulate the neuron firing rate to a basal value.
In these networks, recurrent interactions happen through excitatory/positive synapses, which provide a source of \emph{distributed positive feedback}, and counteract the single-neuron-level negative feedback to create richer and more expressive, but also mathematically tractable and tunable, population-level spiking dynamics.

The problem of determining under which conditions network interactions provide purely positive feedback is at the core of monotone systems theory~\cite{angeli2003monotone}.
Constructive ways to verify or impose such conditions have been summarized in~\cite[Section~II.B]{bizyaeva2023multi}.

\section{From waves to symbols through feedback regulation}
\label{sec:from-waves-symbols}

In this Section, we provide general mechanistic insights into how increasing the strength of the network-level positive feedback is sufficient to shape how the silicon neuron population processes signals and represents information.
In particular, we show a continuous transition from an analog/faithful to a digital/categorical representation and manipulation of incoming inputs.
We use the theory proposed in~\cite{bizyaeva2023multi} to build adjacency matrices $A$ with certified population-level distributed positive feedback.
We analyze the input/output behavior of the resulting mixed-feedback \acp{rSNN} as a function of a positive feedback gain $\alpha\geq0$ that modulates the strength of recurrent interconnections, analogous to the one used in the \ac{sWTA} experiment of Section~\ref{sec:acswta-netw-oper}.
More precisely, given the adjacency matrix $A$, the \ac{rSNN} connectivity is defined as $\alpha A$.

\subsection{Preliminary definitions and notations}
\label{sec:prel-defin-notat}

The key algebraic property of adjacency matrices is that they possess a dominant eigenvalue/eigenvector pair~\cite{bizyaeva2023multi}.
We define them as $\lambda_{max}$ for the eigenvalue, ${\bf v}_{max}$ for the associated right eigenvector and ${\bf w}_{max}$ for the left dominant eigenvector.
The importance of $\lambda_{max}$ is that it determines the critical positive feedback strength at which the more dramatic changes from analog to digital computation take place.
In continuous time, deterministic systems, this change in the system behavior is associated with a bifurcation phenomenon~\cite{leonard2024fast}.
In noisy \acp{rSNN} the same bifurcation explains and predicts the network behaviors and provides a robust design tool.
The relevance of ${\bf v}_{max}$ is that it spans the bifurcation {\it center manifold}. In other words, it predicts the direction in the network state space along which inputs and initial conditions are most strongly and non-linearly amplified.
It can be interpreted as a distributed ultrasensitive state-to-output gain.
In a dual fashion, ${\bf w}_{max}$ predicts the direction in the multi-input space along which the single inputs should co-vary to trigger nonlinear input sensitivity.
Hence,  ${\bf w}_{max}$ can be considered as a distributed ultrasensitive input-to-state gain.
For symmetric adjacency matrices as those considered here, ${\bf w}_{max}={\bf v}_{max}$, but to keep the presented theory general, we will continue to distinguish them in what follows.

We also restrict our analysis to \acp{rSNN} with
$$
A = {\rm diag}({\bf v}_{max})(J - I){\rm diag}({\bf v}_{max}),
$$
where $J\in\mathbb R^{N\times N}$ is the $N$-dimensional matrix of 1s, $I\in\mathbb R^{N\times N}$ is the $N$-dimensional identity matrix, and ${\bf v}_{\max}\in\{-1,1\}^N$ is such that $\sum_{i=1}^N\left({\bf v}_{max}\right)_i=0$.\footnote{For odd $N$, ${\bf v}_{\max}\in\{-1,0,1\}^N$ with one zero entry.}
Hence, $A_{ii}=0$ and $A_{ij}={\rm sign}\left(\left({\bf v}_{\max}\right)_i\left({\bf v}_{\max}\right)_j\right)$ $\forall i,j=1,\ldots,N$, $j\neq i$.
Also note that $A{\bf v}_{max}=\lambda_{max}{\bf v}_{max}$, with $\lambda_{max}={N-1}$, defines the dominant eigenvalue/eigenvector pair.

We consider four output units as defined in Fig.~\ref{fig:sRNNnet} (see also Figs.~\ref{fig:faithful_input_representation}, and~\ref{fig:categorical_representation}) with projection vectors ${\bf v}^{out}_1={\bf v}_{max}$, associated with a symbol ``A'', ${\bf v}^{out}_2=-{\bf v}_{max}$, associated with symbol ``B'', ${\bf v}^{out}_3={\bf v}_{max}^\perp$, associated with symbol ``C'', and ${\bf v}^{out}_4=-{\bf v}_{max}^\perp$, associated with symbol ``D'', where ${\bf v}_{max}^\perp\in\{-1,1\}^N$ is orthogonal to ${\bf v}_{max}$.

Given the \ac{rSNN} output vectors ${\bf v}^{out}_i$, a natural way to summarize the network activity is by computing the output alignment indices:
$$
\rho_i(t)=\frac{{\bf v}^{out}_i}{|| {\bf v}^{out}_i||} \cdot \frac{{\boldsymbol{\nu}}(t) - \bar{\nu}(t)}{||{\boldsymbol{\nu}}(t) ||},\quad \bar\nu(t)=\frac{1}{N}\sum_{i=1}^N \nu_i(t),
$$
where $\boldsymbol{\nu}(t)=[\nu_1(t),\ldots,\nu_N(t)]$ is the vector of network firing rates.
The output alignment index $-1\leq\rho_i(t)\leq 1$ is the cosine similarity between the output projection vector ${\bf v}^{out}_i$ and the centered network firing rate vector ${\boldsymbol{\nu}}(t) - \bar{\nu}(t)$.
It quantifies the strength, or decision, with which the network represents the symbol associated to ${\bf v}^{out}_i$.

Graded changes in the output alignment indexes are indicative of analog signal processing (see Fig.~\ref{fig:faithful_input_representation}), while switching behavior in these indexes are indicative of digital computing (see Fig.~\ref{fig:categorical_representation}).

\subsection{Hardware network structure}
\label{sec:hardw-netw-struct}

To validate the theory we configured the silicon neurons on the neuromorphic chip as in Fig.~\ref{fig:sRNNnet} with a network of $N=8$ excitatory neurons.
As input patterns we considered four stimuli ${\bf v}_1^{in},\ldots,{\bf v}_4^{in}\in\{-1,1\}^N$ aligned with the four output vectors, \emph{i.e.}, ${\bf v}_i^{in}={\bf v}_i^{out}$, and associated with corresponding symbols: ${\bf v}_1^{in}$to~``A'', ${\bf v}_2^{in}$to~``B'', ${\bf v}_3^{in}$to~``C'', ${\bf v}_4^{in}$to~``D''.
Specifically, we defined A=(+1,\,+1,\,+1,\,+1,\,-1,\,-1,\,-1,\,-1), C=(-1,\,+1,\,-1,\,+1,\,-1,\,+1,\,-1,\,+1), and B and D as inverse of A and C respectively.

When applying a specific input stimulus to the network, neuron $i$ receives a spike train with a constant mean rate through an excitatory synapse if the $i$-th component of the input vector associated to the input symbol is $+1$, or through an inhibitory synapse if the $i$-th component of the input vector is $-1$.
The mean spike rate used to stimulate excitatory or inhibitory synapses determines the stimulus strength.

\subsection{Negative feedback promotes analog signal processing}
\label{sec:negat-feedb-prom}

We first explored signal processing and information representation for weak recurrent connections (small $\alpha$) to ensure that negative feedback dominates the network input-output behavior.

\begin{figure}
\centering
\includegraphics[width=0.49\textwidth]{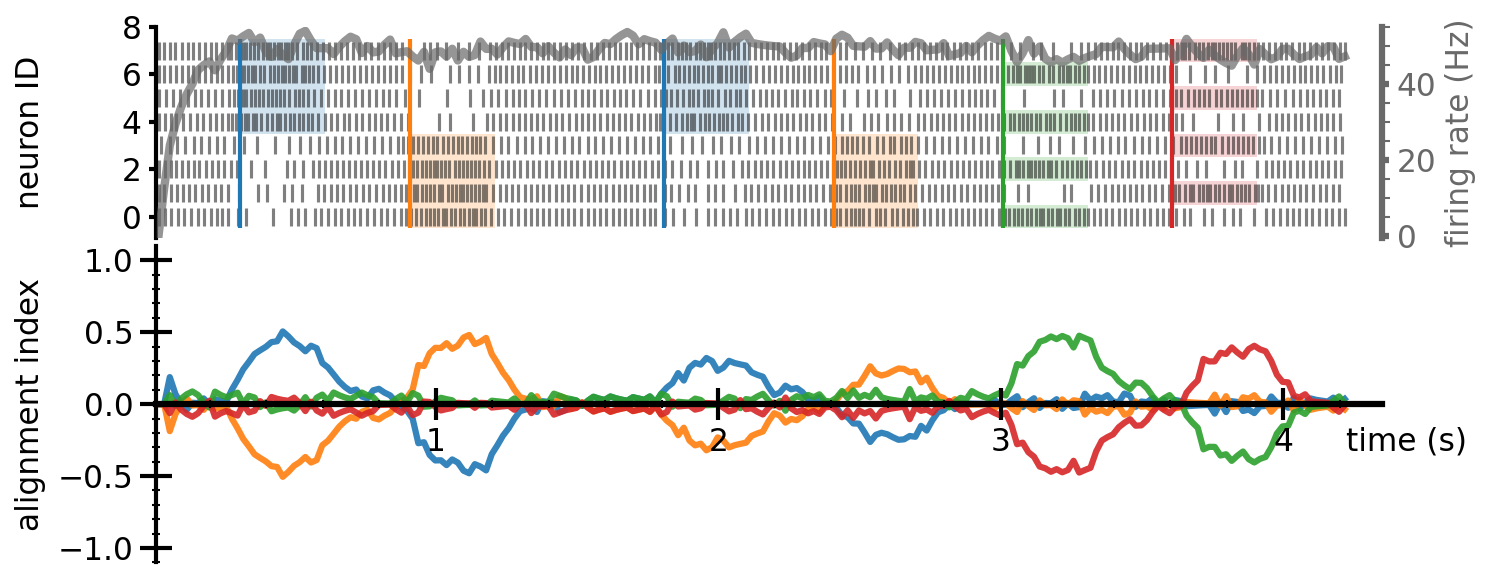}
\caption{Faithful input representation ($\alpha=0.02$).
  First, two inputs A and B (blue and orange) are presented as 200Hz Poisson spike trains through either excitatory or inhibitory synapses (the excitatory inputs are highlighted with the coloured shading on the raster plot, inhibitory ones are omitted for clarity).
  The next two inputs are also A and B, but the firing rates are reduced to 100Hz.
  The last two inputs C and D (green and red) are orthogonal to the clusters.
  The thick gray trace in the top panel represents the network-averaged mean firing rate.
  The lower panel shows the alignment index traces $\rho_1(t) ... \rho_4(t)$ (coloured).}
\label{fig:faithful_input_representation}
\end{figure}

We stimulated the network with a sequence of input symbols (A,\,B,\,A,\,B,\,C,\,D), using corresponding mean spike rates of (200,\,200,\,100,\,100,\,200,\,200)\,Hz, respectively. Each symbol pattern was applied for 300\,ms.
Figure~\ref{fig:faithful_input_representation} shows the measured spiking response of the network with corresponding output alignment indices.
As shown, despite ongoing background noisy activity, each time a stimulus is applied, the network responds by selectively aligning its mean firing rate vector with the output vector associated with the presented symbol.
For instance, when stimulus A is applied, the alignment index $\rho_1$, associated with ${\bf v}_1^{out}={\bf v}_{max}$ grows to about 0.5, the alignment index $\rho_2$, associated with ${\bf v}_2^{out}=-{\bf v}_{max}$, falls to about -0.5, while the alignment indexes $\rho_3,\rho_4$, associated with ${\bf v}_{3}^{out},{\bf v}_{3}^{out}=\pm{\bf v}_{max}^\perp$, remain close to the basal (no stimulus) level of alignment.
The network returns to a noise-driven basal state as soon as stimuli are removed, and weaker stimuli lead to proportionally weaker responses (second repetitions of A,\,B stimuli).
These results confirm that the network tracks its input through negative feedback regulation and behaves as an analog distributed temporal low-pass filter.

\subsection{Strong positive feedback induces the emergence of categorical representations}
\label{sec:strong-pos-feedback}

\begin{figure}
\centering
\includegraphics[width=0.49\textwidth]{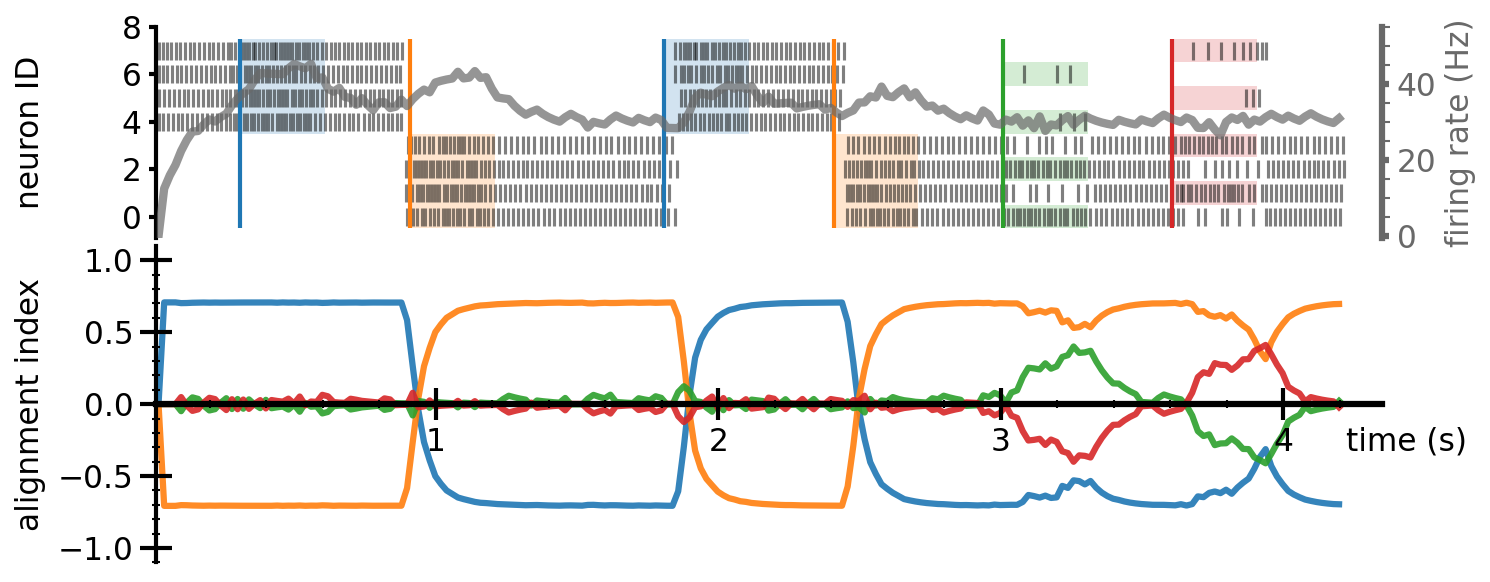}
\caption{Emergence of categorical representations ($\alpha = 0.1$).
Symbols A and B are robustly preserved by the network, even in the case of lower amplitudes of input rates, while symbols C and D do not lead to a persistent state switch after the stimulus is removed.}
\label{fig:categorical_representation}
\end{figure}

In this experiment, we increased the parameter $alpha$ to high values, strengthening all synaptic interconnections in the network.
Figure~\ref{fig:categorical_representation} shows the network response with these new parameter settings.
Consistent with the theory, which predicts the emergence of two exactly two categorical attractors, the network output strongly aligns either with ${\bf v}_{max}$ (corresponding to symbol A in our example) or with $-{\bf v}_{max}$ (corresponding to symbol B).
Hence, as in a binary switch, only two symbols can now be robustly expressed by the network output.
In addition, symbol switching can only robustly be achieved by inputs aligned with ${\bf w}_{max}$\footnote{As a recall, ${\bf w}_{max}$ is the {\it left} dominant eigenvector. In the case considered here ${\bf w}_{max}={\bf v}_{max}$ because the network is undirected and its adjacency matrix is symmetric.}.
This is illustrated by applying the B,A,B stimulus sequence roughly in the middle of Fig.~\ref{fig:categorical_representation}.
Finally, symbol representations are maximally robust to inputs that are orthogonal to ${\bf w}_{max}$.
This is demonstrated by the chip's response to the sequence of C,D stimuli applied at the end of Fig.~\ref{fig:categorical_representation}.

In summary, the network of analog neuron and synapse circuits on the neuromorphic processor works as a distributed digital representation with high-dimensional but tractable and highly tunable bit-switching rules.

\subsection{Robust and smoothly switching from analog to digital using a single parameter}
\label{sec:bifurcation}

According to the theory in~\cite{bizyaeva2023multi}, the switch from analog signal processing to categorical digital computing happens due to the presence of a pitchfork bifurcation in the network state-dynamics as a function of the parameter $\alpha$.
The pitchfork point and the emergence of two symbolic attractors marks the transition from analog processing to digital computing.
Being able to induce this switching behavior using a single parameter allows robust control of the network dynamics around the pitchfork point.

\begin{figure}
\centering
\includegraphics[width=0.4\textwidth]{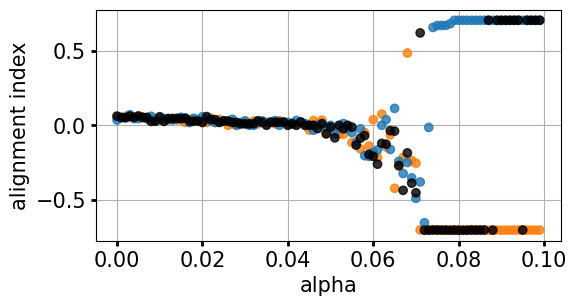}
\caption{Empirical rSNN bifurcation diagram computed using mean firing rates and by projecting the resulting population vector of steady-state rates onto the dominant eigenvector ${\bf v}_{max}$. The colours indicate different initial conditions before the rate estimation is taken: stimulus A (blue), stimulus B (orange) or only background DC input (black). Each rate estimation was performed after the removal of the stimulus.}
\label{fig:bifurcation_diagram}
\end{figure}
We verified the robust existence of this pitchfork point experimentally on the hardware by calculating the empirical bifurcation diagram of the \ac{rSNN} from the mean firing rates of the silicon neurons measured in response to the same inputs for different values of $\alpha$.
Figure~\ref{fig:bifurcation_diagram} shows the data obtained from these measurements.
In line with the theory, the critical $\alpha$ value at which the bifurcation happens is around 0.065.

\,
These results confirm that\,
These results confirm that\section{Discussion and conclusion}
\label{sec:discussion}

The notion that cortical circuits exhibit both analog and digital computational properties has already been made evident in multiple experimental and theoretical studies~\cite{Douglas_etal95,Jaeger99,VanRullen_Koch03,Rutishauser_etal15,Jaeger_etal23}.
In this paper we demonstrated that spiking neural networks implemented with neuromorphic electronic circuits designed to emulate the physics of biological neurons can also exhibit this dualism, using the same computational substrate.
We first showed this by configuring the neuromorphic chip to implement a spiking neural network with structure and connectivity properties similar to those of cortical microcircuits~\cite{Douglas_Martin07} (namely a \ac{sWTA} architecture). \
We showed how it was possible to make the behavior of this network switch from analog processing to categorical non-linear processing by modulating a single parameter.
Then we developed a formal theoretical framework to characterize quantitatively these switching properties for a more generic \ac{rSNN} network configuration and validated the theoretical predictions experimentally with the same chip.
For the sake of clarity and conciseness, we restricted the theoretical analysis presented in Section~\ref{sec:from-waves-symbols} to signed but unweighted dominant eigenvectors and resulting adjacency matrices and symbolic attractors.
However, the same theory accommodates graded dominant eigenvectors, which would lead to symbolic attractors represented by graded network activity and with graded symbol-switching rules.
This would constitute a theoretically grounded generalization of the sWTA behavior to arbitrary graded symbols and enable the robust control of electronic devices and systems that use analog components, even if affected by device mismatch and noise~\cite{Zendrikov_etal23}.

As mixed-signal neuromorphic circuits dissipate very low power, in the pico- to micro-Watt range~\cite{Rubino_etal20}, they represent a very promising technology for  use in sensors and sensory-processing systems that have severe constraints on resources such as memory and power consumption (e.g., in edge-computing applications).
The work presented here will allow to endow such signal processing systems also with robust computing abilities, without having to resort to power-hungry digital processing systems.

\renewcommand{\bibfont}{\footnotesize} 
\printbibliography
\end{document}